\documentclass[a4paper,conference]{IEEEtran}
\usepackage{cite}
\usepackage{amsmath,amssymb,amsfonts}
\usepackage{algorithmic}
\usepackage{graphicx}
\usepackage{textcomp}
\usepackage{xcolor}

\usepackage{times}
\usepackage{epsfig}
\usepackage{caption}
\usepackage{subfigure}
\usepackage{threeparttable}
\usepackage[marginal, flushmargin]{footmisc}
\usepackage{algorithm}
\usepackage{algorithmic}
\usepackage{multirow}
\usepackage{bm}
\usepackage{mathrsfs}
\usepackage{url}

\usepackage{comment}
\usepackage{color}
\usepackage{etoolbox}
\newbool{inccomment}
\setbool{inccomment}{true}

\newcommand{\tabincell}[2]{\begin{tabular}{@{}#1@{}}#2\end{tabular}}

\begin{document}

\title{Conditional Transferring Features:\\ Scaling GANs to Thousands of Classes with 30\% Less High-quality Data for Training}

\author{\IEEEauthorblockN{Chunpeng Wu, Wei Wen, Yiran Chen, and Hai Li}
\IEEEauthorblockA{Department of Electrical and Computer Engineering, Duke University, Durham, NC 27708\\
\{chunpeng.wu, wei.wen, yiran.chen, hai.li\}duke.edu}}

\maketitle

\begin{abstract}
Generative adversarial network (GAN) has greatly improved the quality of unsupervised image generation.
Previous GAN-based methods often require a large amount of high-quality training data while producing a small number (\textit{e.g.}, tens) of classes.
This work aims to scale up GANs to thousands of classes meanwhile reducing the use of high-quality data in training.
We propose an image generation method based on conditional transferring features,  which can capture pixel-level semantic changes when transforming low-quality images into high-quality ones.
Moreover, self-supervision learning is integrated into our GAN architecture to provide more label-free semantic supervisory information observed from the training data.
As such, training our GAN architecture requires much fewer high-quality images with a small number of additional low-quality images.
The experiments on CIFAR-10 and STL-10 show that even removing 30\% high-quality images from the training set, our method can still outperform previous ones.
The scalability on object classes has been experimentally validated:
our method with 30\% fewer high-quality images obtains the best quality in generating 1,000 ImageNet classes, as well as generating all 3,755 classes of CASIA-HWDB1.0 Chinese handwriting characters.
\end{abstract}

\IEEEpeerreviewmaketitle

\section{Introduction}
\label{sec:introduction}

As one of the most exciting breakthroughs in unsupervised machine learning, \textit{generative adversarial network} (GAN)~\cite{Goodfellow_NIPS14} has been successfully applied to a variety of applications, such as face verification~\cite{Li_AAAI18}, human pose estimation~\cite{Chen_ICCV17} and small object detection~\cite{Li_CVPR17}. 
In principle, GANs are trained in an adversarial manner: a \textit{generator} produces new data by mimicking a targeted distribution; meanwhile, a \textit{discriminator} measures the similarity between the generated and targeted distributions, which in turn is used to adapt the generator. 

\begin{figure}[t!]
	\centering
	\includegraphics[width=7cm]{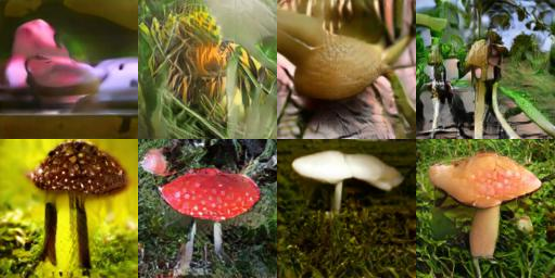}
    \caption{Our experiments on GAN-based image generation.The top and bottom rows are generated mushroom images by using 60\% of and 100\% of the ImageNet training set, respectively.}
    \label{fig:cmp_img_gen_100_60}
\vspace{-15pt}
\end{figure}

The quality of generated data highly relies on both \textit{volume} and \textit{quality} of the training data. 
For example, our experiments on GAN-based image generation and image-to-image translation show dramatic performance degradation when reducing the number of high-quality training images.
Figure~\ref{fig:cmp_img_gen_100_60} shows several mushroom images generated by SN-GAN~\cite{Miyato_ICLR18} trained with 60\% of (top row) or 100\% of (bottom row) ImageNet training data~\cite{Russakovsky_IJCV15}. 
The images in the bottom row obtained by using the entire training dataset present more distinguishable appearance (\textit{e.g.}, cap and stem of mushroom) and have much better quality. 
After removing 40\% of the training data, the Inception score decreases from 21.1 to 14.8 and FID increases from 90.4 to 141.2.
\textit{The high demand on high-quality training data has emerged as a major challenge of GAN-based methods---it is very difficult or even impossible to collect sufficient data for producing satisfactory results in real-world applications.}

To address the challenge, we propose an image generation method based on \textit{conditional transferring features} (CTFs) with three key solutions.
First, we contruct the training data with a portion of the original high-quality images and a small number of low-quality images. 
Second, our method extracts the CTFs by transforming low-quality images into high-quality images.
Third, we further enhance our method with more label-free supervisory information observed from the training data.

\section{Related Work}
\label{sec:related_work}

\textbf{Many GAN research studies} explore how to stabilize GAN training through modifying network architecture~\cite{Radford_ICLR16,Gulrajani_NIPS17} and optimizing algorithms~\cite{Arjovsky_arxiv17,Mao_ICCV17}.
The recent SN-GAN~\cite{Miyato_ICLR18} stabilizes the discriminator of a GAN using a novel weight normalization method.
\textit{In this work, we propose a new approach rarely considered in GAN-based methods, that is, using low-quality training data to facilitate the generation of high-quality images.}

\begin{figure}[t]
	\centering
	\includegraphics[width=\columnwidth]{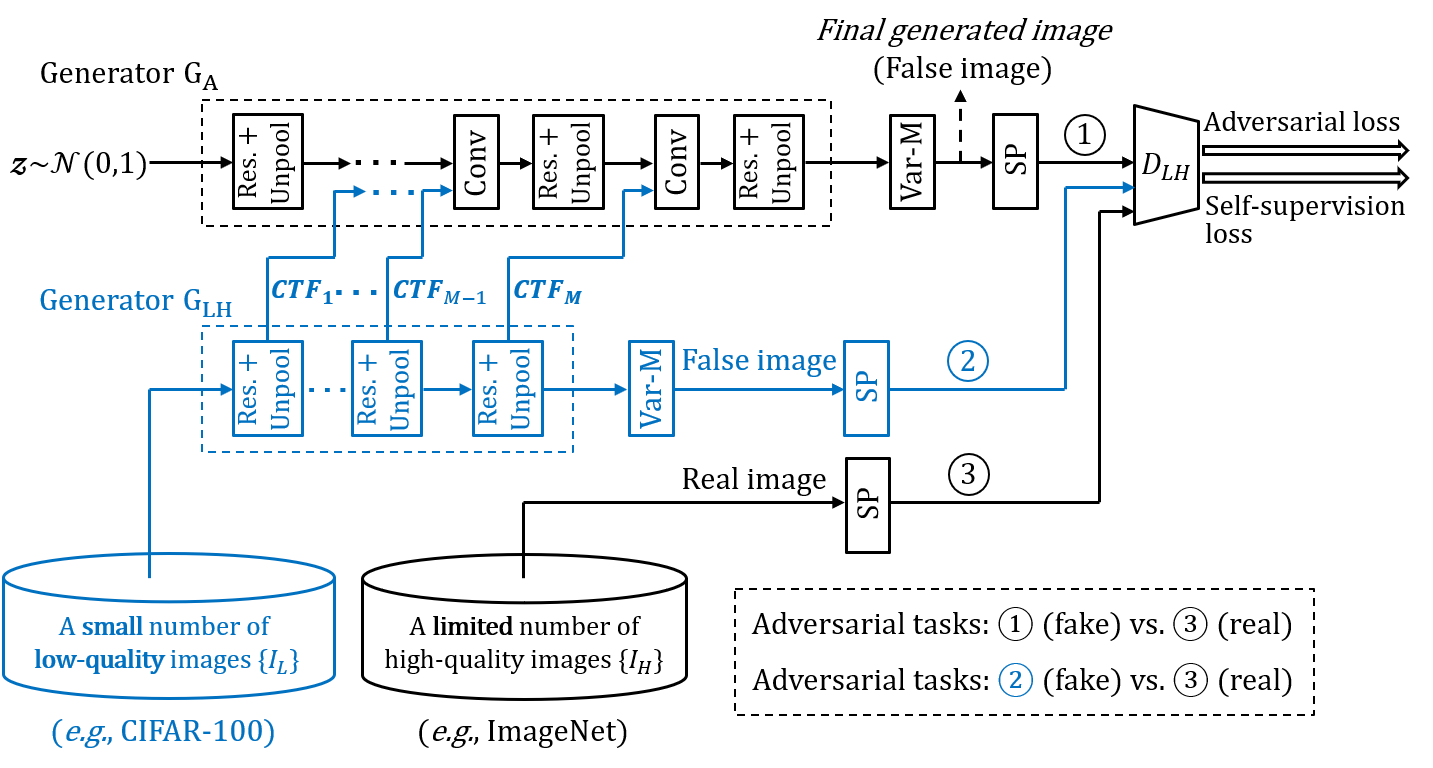}
	\caption{Our proposed image generation method. The data-flow in blue extracts our proposed CTFs and provides CTFs to generator $G_{A}$. Tasks \raisebox{.5pt}{\textcircled{\raisebox{-.9pt} {2}}} (in blue) and \raisebox{.5pt}{\textcircled{\raisebox{-.9pt} {3}}} (in dotted lines) are first adversarially trained, followed by the training of adversarial tasks \raisebox{.5pt}{\textcircled{\raisebox{-.9pt} {1}}} and \raisebox{.5pt}{\textcircled{\raisebox{-.9pt} {3}}}. The number of high-quality images $I_{H}$ and low-quality images $I_{L}$ are not required to be the same.}
	\label{fig:img_gen_whole_framework}
\vspace{-15pt}
\end{figure}

\textbf{GAN-based image generation methods~\cite{Denton_NIPS15,Salimans_NIPS16_2,Durugkar_ICLR17,Dumoulin_ICLR17,Dai_ICLR17,Yang_ICLR17,Warde-Farley_ICLR17,Grinblat_arxiv17,Salimans_NIPS16,Heusel_NIPS17,Odena_ICML17,Karras_ICLR18}} tackle the issues of multi-resolution, variation observation, architecture changing, energy estimation for samples, embedding recursive structures, integrating condition information into GANs, and quality evaluation of generated images.
Recently, BigGAN~\cite{Brock_ICLR19} significantly improves image synthesis quality by adding orthogonal regularization to the generator.
\textit{Our GAN-based method can scale to thousands of classes with significantly fewer high-quality training data.}

\section{Image Generation Based on CTFs}
\label{sec:our_gen}

Figure~\ref{fig:img_gen_whole_framework} shows our proposed image generation method.
Following traditional GAN-based methods, the design consists of a generator $G_{A}$ and a discriminator $D_{LH}$.
Our discriminator is also used for \textit{self-supervision} (SP) learning.
We introduce a generator $G_{LH}$ for extracting CTFs.
There are three learning tasks in our method:
\begin{itemize}
\item Task \raisebox{.5pt}{\textcircled{\raisebox{-.9pt} {1}}} adopts $G_{A}$ and $D_{LH}$ to produce images similar to the high-quality $I_{H}$ by using noise $z_{A}\sim \mathcal{N}(0,1)$ and the conditional transferring features $\bm{CTF_{m}}\,(m=1,2,...,M)$ as $G_{A}$'s input.
The \textit{Conv} layers in $G_{A}$ convolute $\bm{CTF_{m}}\, (m=1,2,...,M)$ by taking the output from the previous layer under the same resolution.
\item Task \raisebox{.5pt}{\textcircled{\raisebox{-.9pt} {2}}} (highlight in blue) adopts $G_{LH}$ and $D_{LH}$ to transform the low-quality images to high-quality images similar to $I_{H}$ and provides the extracted $\bm{CTF_{m}}\,(m=1,2,...,M)$ to $G_{A}$. 
Noises $z_m\,(m=1,2,...,M)$ are injected into each \textit{Res.+Unpool} (\textit{ResBlock+Unpooling}) block in $G_{LH}$, respectively, to increase the randomness of the generated images.
\item Task \raisebox{.5pt}{\textcircled{\raisebox{-.9pt} {3}}} (represented in dotted lines) is to distinguish the real images sampled $I_{H}$ from synthetic images using $D_{LH}$.
\end{itemize}

\noindent
The adversarial tasks \raisebox{.5pt}{\textcircled{\raisebox{-.9pt} {2}}} and \raisebox{.5pt}{\textcircled{\raisebox{-.9pt} {3}}} are first trained for extracting the CTFs until no significant improvement can be observed.
Afterwards, tasks \raisebox{.5pt}{\textcircled{\raisebox{-.9pt} {1}}} and \raisebox{.5pt}{\textcircled{\raisebox{-.9pt} {3}}} are adversarially trained for image generation based on the CTFs.

\begin{figure}[t]
	\centering
    \includegraphics[width=\columnwidth]{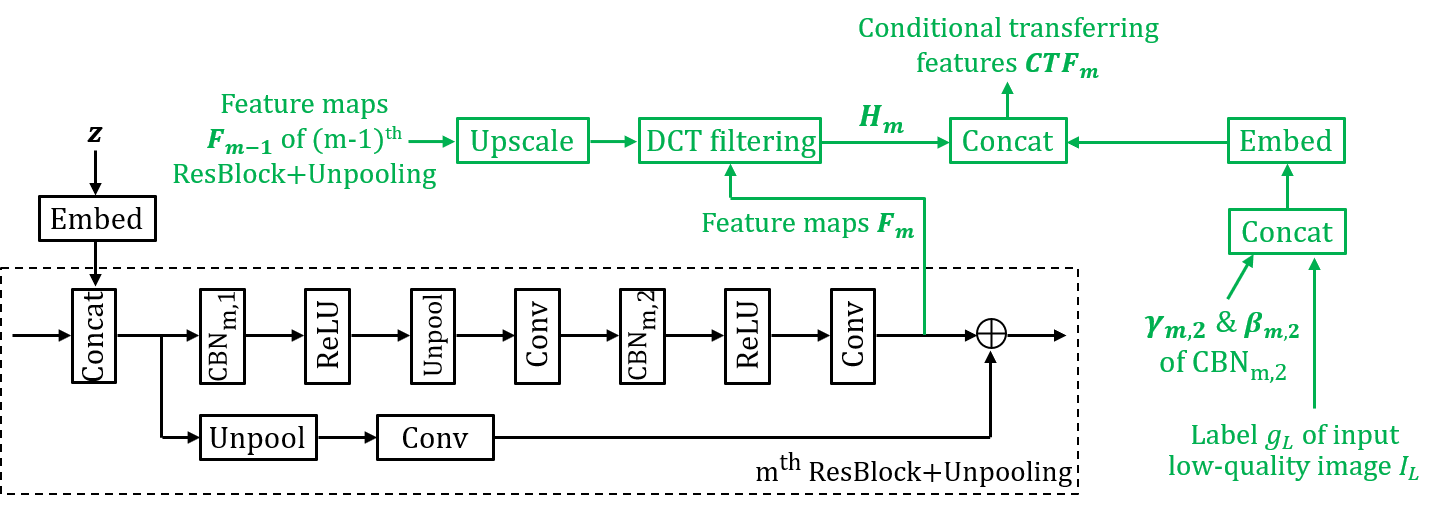}
	\caption{Data-flow (in green) for extracting the conditional transferring features $\bm{CTF_{m}}$ in $m^{th}$ \textit{ResBlock+Unpooling} block in the generator $G_{LH}$.}
	\label{fig:our_resblock}
\vspace{-15pt}
\end{figure}

\subsection{Extracting CTFs}

In $G_{LH}$, the $m^{th}$ \textit{ResBlock+Unpooling} block is used to extract the conditional transferring feature $\bm{CTF_{m}}$.
Figure~\ref{fig:our_resblock} shows the detailed extraction data-flow (in green).
The random noise $z_{m}\sim \mathcal{N}(0,1)$ is embedded and concatenated to the input of the block. 
The $\mathrm{Embed}$ operator is previously described in~\cite{Reed_ICML16,Miyato_ICLR18,Miyato_ICLR18_2}.
We replace the batch-normalization~\cite{Ioffe_arxiv15} layers in traditional ResBlock with \textit{conditional batch-normalization} (CBN) \cite{Dumoulin_ICLR17_2} layers $CBN_{m,1}$ and $CBN_{m,2}$ in \textit{ResBlock+Unpooling}.
$CBN_{m,1}$ and $CBN_{m,2}$ are conditional to the label information $c\in \{1,...,c_{H}\}$ of the high-quality images where $c_{H}$ is the class number of the high-quality images.
According to the CBN's definition~\cite{Dumoulin_ICLR17_2}, for the layer $CBN_{m,1}$, an input activation $x_{m,1}$ is transformed into a normalized activation $y_{m,1}$ specific to a class $c\in \{1,...,c_{H}\}$ calculated as:
\begin{equation}
\label{eq:cbn_1}
y_{m,1}=\gamma_{m,1}^{c}\frac{x_{m,1}-\mu}{\sigma}+\beta_{m,1}^{c} \, ,
\end{equation} 
where $\mu$ and $\sigma$ are respectively the mean and standard deviation taken across spatial axes, and $\gamma_{m,1}^{c}$ and $\beta_{m,1}^{c}$ are trainable parameters specific to class $c$ of $CBN_{m,1}$.
Thus, the trainable parameters of $CBN_{m,1}$ are $\bm{\gamma_{m,1}}=\{\gamma_{m,1}^{c}\}_{c=1}^{C_{H}}$ and $\bm{\beta_{m,1}}=\{\beta_{m,1}^{c}\}_{c=1}^{C_{H}}$. 
Similarly, $\bm{\gamma_{m,2}}=\{\gamma_{m,2}^{c}\}_{c=1}^{C_{H}}$ and $\bm{\beta_{m,2}}=\{\beta_{m,2}^{c}\}_{c=1}^{C_{H}}$ denote the trainable parameters across all the classes of $CBN_{m,2}$.

The label information of both low-quality and high-quality images are concatenated to feature the differences between adjacent blocks of \textit{ResBlock+Unpooling}.  
$\bm{CTF_{m}}$ is calculated by:
\small
\begin{equation}
\label{eq:ctf}
    \begin{aligned}
        \bm{CTF_{m}}={\rm Concat}\Big(\bm{H_{m}}, \, {\rm Embed}\big({\rm Concat}(\bm{\gamma_{m,2}}, \bm{\beta_{m,2}}, \bm{g_{L}} )\big)\Big)\\
    \end{aligned},
\end{equation} 
\normalsize
where $\bm{H_{m}}=\{H_{m}^{t}\}_{t=1}^{T}$ is the aggregated difference maps between the feature maps $\bm{F_{m}}=\{F_{m}^{t}\}_{t=1}^{T}$ and $\bm{F_{m-1}}=\{F_{m-1}^{s}\}_{s=1}^{S}$ respectively in the $m^{th}$ and the $(m-1)^{th}$ blocks of \textit{ResBlock+Unpooling}. 
Note that $T$ might not be equal to $S$.
To make it more clear, given a feature map $F_m^{t}$, the difference map between $F_m^{t}$ and each $F_{m-1}^{s}\, (s=1,2,...,S)$ is calculated, and then $H_{m}^{t}$ is obtained by aggregating all $S$ difference maps together.
$\bm{g_{L}}$ is the labels of input low-quality images $I_{L}$, and $\bm{\gamma_{m,2}}$ and $\bm{\beta_{m,2}}$ include label information of high-quality images.
The class information of low-quality and high-quality images are first concatenated together before they are concatenated to the difference maps $\bm{H_{m}}$.
The class conditional parameters $\bm{\gamma_{m,1}}$ and $\bm{\beta_{m,1}}$ of the layer $CBN_{m,1}$ are not used in Equation~(\ref{eq:ctf}) because the layer $CBN_{m,2}$ is in front of the \textit{Unpooling} layer as shown in Figure~\ref{fig:our_resblock}, \textit{i.e.}, its resolution corresponds to $\bm{F_{m-1}}$ but not $\bm{F_{m}}$.
The feature maps $\bm{F_{m-1}}$ will be upsampled to the same size of $\bm{F_{m}}$ using bilinear interpolation.
For the first block of \textit{ResBlock+Unpooling} ($m=1$), the previous $\bm{F_{m-1}}$ is replaced by the gray-level version of low-quality image $I_{L}$.
The differences between a pair of feature maps are evaluated in a DCT-based frequency domain $\mathscr{D}$.
$H_{m}^{t}\, (t=1,2,...,T)$ is calculated as shown in Equation~(\ref{eq:fft_filter_1}) when $m=1$:
\small
\begin{equation}
	\label{eq:fft_filter_1}
    H_{m}^{t}=\mathscr{D}^{-1}\Big (\mathscr{D}(F_{m}^{t}) - \mathscr{D}({\rm Upscale}({\rm CvtGray}(I_{L})) ) \Big ),
\end{equation}
\normalsize
where $\mathscr{D}(\cdot)$ and $\mathscr{D}^{-1}(\cdot)$ are DCT and inverse DCT transforms, ${\rm Upscale}$ function unsamples an image using bilinear interpolation, and ${\rm CvtGray}$ function converts a color image into a gray-level image.
$H_{m}^{t}\, (t=1,2,...,T)$ is calculated as shown in Equation~(\ref{eq:fft_filter_2}) when $1 < m \leq M$:
\small
\begin{equation}
    \label{eq:fft_filter_2}
    H_{m}^{t}=\mathscr{D}^{-1}\left (\frac{\sum_{s=1}^{S} (\mathscr{D}(F_{m}^{t}) -  \mathscr{D}({\rm Upscale}(F_{m-1}^{s}))  )}{S} \right ).
\end{equation}
\normalsize

\subsection{Self-supervision (SP) learning}

Popular self-supervision tasks~\cite{Dosovitskiy_NIPS14,Gidaris_ICLR18,Noroozi_ECCV16} predict chromatic transformations, rotation, scaling, relative position of the image patches, \textit{etc}.
Our self-learning method is inspired by~\cite{Dwibedi_ICCV17}.
For any image sampled from $I_{H}$ or generated by $G_{A}$ (or $G_{LH}$), we randomly cut an image patch from $I_{H}$, paste it to a random location of the image, and record the bounding-box coordinates of the patch.
The \textit{Self-supervision loss} shown in Figure~\ref{fig:img_gen_whole_framework} is the differences between the recorded coordinates and those predicted by the discriminator $D_{LH}$.

\setlength{\tabcolsep}{10pt}
\begin{table*}
	\begin{center}
		\caption{Conditional image generation on CIFAR-10 and STL-10.}
		\label{table:exp_cifar10_stl10}
		\small
\begin{threeparttable}
	\begin{tabular}{l|c|c|c|c}
		\hline\noalign{\smallskip}
		\multirow{2}{*}{Method} & \multicolumn{2}{c|}{CIFAR-10} & \multicolumn{2}{c}{STL-10}\\
		& Training data & Inception score & Training data & Inception score \\
		\hline\hline\noalign{\smallskip}		
		AC-GAN~\cite{Odena_ICML17}              & \multirow{3}{*}{\tabincell{c}{50,000\\(CIFAR-10\\training set)}} & $8.25\pm .07$ & \multirow{3}{*}{\tabincell{c}{5,000\\(STL-10\\training set)}} & - \\
		PROG-GAN~\cite{Karras_ICLR18}           & & $8.88\pm .05$ &  & $9.34\pm .06$ \\
		SN-GAN+Projection                       & & $9.01\pm .04$ &  & $9.38\pm .08$ \\
		\hline\noalign{\smallskip}
		Ours (Fewer HQ)               & \textbf{32,500+10,000} & $9.05\pm .06$ & \textbf{3,250+1,000} & $9.44\pm .04$ \\
		\textbf{Ours (Entire HQ)}     & 50,000+10,000 & $\bm{9.17\pm .04}$ & 5,000+1,000 & $\bm{9.63\pm .05}$\\	
		\hline
	\end{tabular}
\end{threeparttable}
	\end{center}
\vspace{-15pt}
\end{table*}
\setlength{\tabcolsep}{10pt}

\setlength{\tabcolsep}{1pt}
\begin{table}[t]
	\begin{center}
		\caption{Conditional image generation on CASIA-HWDB1.0.}
		\label{table:exp_chn}
		\small
		\begin{threeparttable}
			\begin{tabular}{l|c|c}
				\hline\noalign{\smallskip}
				Method & Training data & Inception score\\
				\hline\hline\noalign{\smallskip}
				SN-GAN+Projection            & 1,246,991 & $10.2 \pm .17$ \\
				Ours (Fewer HQ)              & \textbf{880,544 (810,544+70,000)} & $11.3 \pm .13$ \\
				\textbf{Ours (Entire HQ)}    & 1,316,991 (1,246,991+70,000) & $\bm{13.6 \pm .15}$ \\                	
				\hline
			\end{tabular}
		\end{threeparttable}
	\end{center}
\end{table}
\setlength{\tabcolsep}{1pt}

\section{Experiments}
\label{sec:experiments}

\setlength{\tabcolsep}{2pt}
\begin{table}[t]
	\begin{center}
		\caption{Conditional image generation on ImageNet.}
		\label{table:exp_imagenet}
		\small
		\begin{threeparttable}
			\begin{tabular}{l|c|c}
				\hline\noalign{\smallskip}
				Method & Training data & Inception score\\
				\hline\hline\noalign{\smallskip}
				AC-GAN~\cite{Dumoulin_ICLR17}  & \multirow{3}{*}{\tabincell{c}{1,282,167\\(ImageNet training set)}} & $28.5\pm .20$\\
				SN-GAN~\cite{Miyato_ICLR18}                      & & $21.1\pm .35$\\
				SN-GAN+Projection           & & $36.8\pm .44$\\
				\hline\noalign{\smallskip}
				Ours (Fewer HQ)               & \textbf{893,408 (833,408+60,000)} & $39.4\pm .43$\\
				\textbf{Ours (Entire HQ)}     & 1,342,167 (1,282,167+60,000) & $\bm{43.9\pm .46}$\\                   	
				\hline
			\end{tabular}
		\end{threeparttable}
	\end{center}
	\vspace{-10pt}
\end{table}
\setlength{\tabcolsep}{2pt}

Our generator $G_{A}$ has similar architecture with the generator adopted in SN-GAN+Projection~\cite{Miyato_ICLR18,Miyato_ICLR18_2}, but two differences are as follows.
Our $G_{A}$ has additional layers for convoluting with CTFs provided by the generator $G_{LH}$.
Our $G_{A}$ uses regular BN (instead of CBN in SN-GAN+Projection), while our $G_{A}$ uses CBN.

\subsection{Conditional Image Generation on CIFAR-10 and STL-10}
\label{sec:exp_cifar10_stl10}

Table~\ref{table:exp_cifar10_stl10} compares the quality of image generation by using previous methods~\cite{Radford_ICLR16,Salimans_NIPS16_2,Odena_ICML17,Huang_CVPR17,Gulrajani_NIPS17,Grinblat_arxiv17,Karras_ICLR18,Miyato_ICLR18,Miyato_ICLR18_2} with ours. 
All the previous approaches take the entire CIFAR-10 training set (50,000 images).
Our training data is a mixed-up of \textit{high-quality} (HQ) images sampled from CIFAR-10 or STL-10 training set and \textit{low-quality} (LQ) images.
Since CIFAR-10 or STL-10 are already the ``simplest'' datasets, we have to use their down-sampled versions as LQ images.
For comparison purpose, \textit{Ours (Fewer HQ)} uses 32,500 CIFAR-10 and 10,000 down-sampled images as training data, and \textit{Ours (Entire HQ)} applies the entire CIFAR-10 training set and 10,000 down-sampled images.
According to the popular testing protocol~\cite{Gulrajani_NIPS17,Grinblat_arxiv17,Miyato_ICLR18}, we scale all the generated images to 32$\times$32 for CIFAR-10 classes and 48$\times$48 for STL-10 classes.

The experiment shows that \textit{Ours (Fewer HQ)} with fewer training data slightly outperforms previous methods.
Using the entire CIFAR-10 or STL-10 training sets further improves the image quality of our method: 
\textit{Ours (Entire HQ)} is respectively 1.7\% and 2.7\% better in Inception score, compared to previously best SN-GAN+Projection~\cite{Miyato_ICLR18,Miyato_ICLR18_2}.

\subsection{Conditional Image Generation on 3755-Class CASIA-HSWB1.0}

To further validate the scalability on object classes, we compare the generation of 3,755 classes of CASIA- HWDB1.0 Chinese characters by using our method and SN-GAN+Projection~\cite{Miyato_ICLR18,Miyato_ICLR18_2}.
SN-GAN+Projection adopts the entire CASIA-HWDB1.0 training set (1,246,991 images) as training data.
Our training data takes 810,544 CASIA-HWDB1.0 training set as HQ images and 70,000 MNIST handwriting images as LQ images.
The total number of our training data is (880,544) is 29.4\% smaller than the entire CASIA-HWDB1.0 training set.
The resolution of the generated images is set to 48$\times$48 which is the same as original CASIA-HWDB1.0 dataset.

The quantitative comparison in Table~\ref{table:exp_chn} validates that \textit{Ours (Fewer HQ)} and \textit{Ours (Entire HQ)} can produce higher-quality Chinese characters in 3,755 CASIA-HWDB1.0 classes, compared to SN-GAN+Projection.
The quality gap between SN-GAN+Projection and \textit{Ours (Fewer HQ)} is larger than the gap presented in Table~\ref{table:exp_cifar10_stl10}, which implies our advantage on more image classes.

\begin{figure}[t!]
	\centering
	\includegraphics[width=0.8\columnwidth]{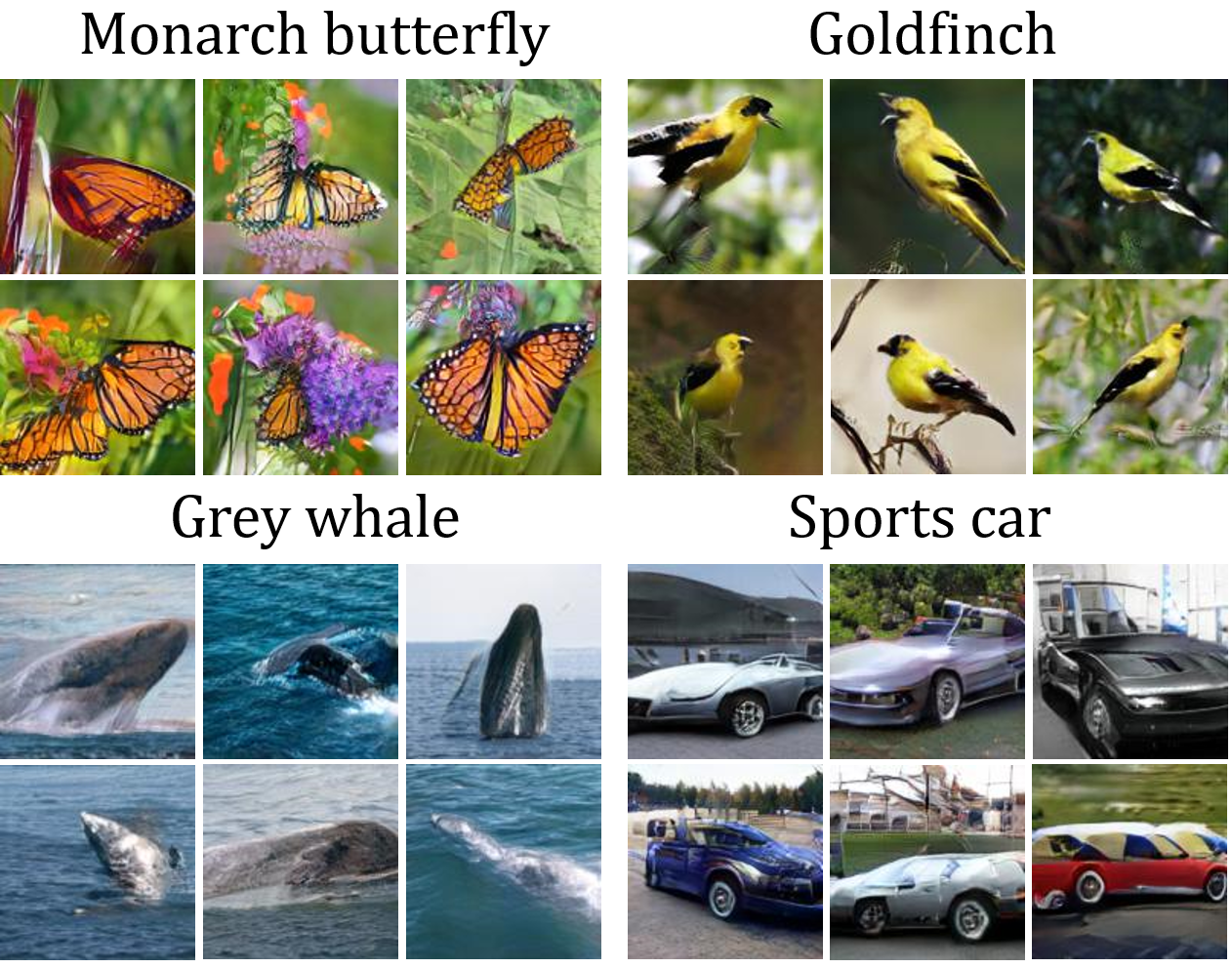}
	\caption{Our generated images of the ImageNet classes (\textit{Ours (Entire HQ)}).}
	\label{fig:exp_gen_imagenet}
	\vspace{-5pt}
\end{figure}

\subsection{Conditional Image Generation on ImageNet}
\label{sec:exp_imagenet}

We use our method for conditional image generation on ImageNet classes and compare it to AC-GAN~\cite{Dumoulin_ICLR17}, SN-GAN~\cite{Miyato_ICLR18} and SN-GAN+Projection~\cite{Miyato_ICLR18,Miyato_ICLR18_2}. 
The training of the three previous GANs adopt the entire ImageNet training set (1,282,167 images).
The training data of \textit{Ours (Fewer HQ)} contains 833,408 ImageNet training set as HQ images and 60,000 CIFAR-100 images as LQ ones.
Thus, the total number of \textit{Ours (Fewer HQ)} is 30.3\% smaller than the entire ImageNet training set used in previous methods~\cite{Odena_ICML17,Miyato_ICLR18,Miyato_ICLR18_2}.
The resolution of the generated images is set to 128$\times$128 to compare with previous methods.

Table~\ref{table:exp_imagenet} summarizes the comparison with previous methods, and Figure~\ref{fig:exp_gen_imagenet} shows some examples of the generated images by \textit{Ours (Entire HQ)}.
\textit{Ours (Fewer HQ)} outperforms previous methods, even though it uses 30.3\% fewer training data.
Using the entire ImageNet training set and CIFAR-100 images, \textit{Ours (Entire HQ)} is 17.0\% better than the previous best SN-GAN+Projection \cite{Miyato_ICLR18,Miyato_ICLR18_2} in Inception score.

\section{Conclusion}
\label{sec:conclusion}
Previous GAN-based image generation methods face the challenges of the heavy dependency on high-quality training data.
In contrast, collecting low-quality images is relatively easier and more economical.
Through the observation on the learning process during transforming the low-quality images into high-quality images, we find that certain intermediate output combined with the class information, or \textit{conditional transferring features} (CTFs), can be adopted to improve the quality of image generation and scalability on object classes of GAN.
Moreover, we integrate self-supervision learning into our GAN architecture to further improve the learning ability of the GAN.
Experiments on conditional image generation tasks show that our method performs better than previous methods even when 30\% high-quality training data is removed.
And our method successfully scales GANs to thousands of object classes such as the 1,000 ImageNet classes and 3,755 CASIA-HWDB1.0 classes.


{\small
\bibliographystyle{IEEEtran}
\bibliography{IEEEabrv,egbib}
}

\end{document}